\pdfoutput=1
\documentclass[conference,a4paper]{IEEEtran}
\IEEEoverridecommandlockouts

\IEEEsettextheight{0.75in}{1in}
\IEEEquantizetextheight{c}

\usepackage{cite}

\usepackage{amsmath,amssymb,amsfonts}
\usepackage{algorithmic}
\usepackage{graphicx}
\usepackage{textcomp}
\usepackage{xcolor}
\def\BibTeX{{\rm B\kern-.05em{\sc i\kern-.025em b}\kern-.08em
    T\kern-.1667em\lower.7ex\hbox{E}\kern-.125emX}}

\usepackage{booktabs}
\usepackage{multirow}
\usepackage{url}
\newcommand{\codelink}[2]{\pdfstartlink attr{/Border[0 0 0]} user{/Subtype/Link/A<</Type/Action/S/URI/URI(#1)>>}#2\pdfendlink}
\newcommand{\second}[1]{\textcolor{blue}{#1}}

\makeatletter
\g@addto@macro\itemize{\setlength{\itemsep}{1pt}\setlength{\parsep}{0pt}}
\g@addto@macro\enumerate{\setlength{\itemsep}{1pt}\setlength{\parsep}{0pt}}
\makeatother

\begin{document}
\bstctlcite{IEEEexample:BSTcontrol}

\title{Neural Image Compression with Internal-External Prior Fusion and Alignment}

\author{\IEEEauthorblockN{Tiancheng Si}
\IEEEauthorblockA{\textit{School of Artificial Intelligence} \\
\textit{Xidian University}\\
Xi'an 710126, China \\
tiancheng@stu.xidian.edu.cn}
\and
\IEEEauthorblockN{Wenlong Xu}
\IEEEauthorblockA{\textit{School of Artificial Intelligence} \\
\textit{Xidian University}\\
Xi'an 710126, China \\
xuwl@stu.xidian.edu.cn}
\and
\IEEEauthorblockN{Dahua Gao}
\IEEEauthorblockA{\textit{School of Artificial Intelligence} \\
\textit{Xidian University}\\
Xi'an 710126, China \\
dhgao@xidian.edu.cn}%
\thanks{\textit{Corresponding author: Dahua Gao.}}
}

\IEEEaftertitletext{\vspace{-1.25\baselineskip}}
\maketitle

\begin{abstract}
Existing context models for learned image compression rely primarily on internal priors from the current image, overlooking common patterns and typical textures shared across natural images. To address this limitation, we propose an Internal-External Prior Fusion-Alignment (IE-PFA) context model built upon quadtree spatial-channel joint modeling, where a learnable dictionary shared by the encoder and decoder serves as an external prior. Specifically, an RWKV-based Spatial-Channel Prior Fusion (RSCPF) module efficiently fuses internal and external priors along spatial and channel dimensions with linear complexity, while a training-only Self-Distilled Prior Alignment (SDPA) mechanism aligns them without inference overhead. Moreover, our model supports variable-rate coding over 64 quality levels within a single network and is accelerated by custom CUDA operators. Experiments on seven benchmarks demonstrate state-of-the-art rate-distortion performance, achieving an average BD-Rate saving of 21.65\% over VTM-22.0-Intra, with only 29.9M parameters and the fastest coding speed among advanced neural codecs (61 ms encoding and 52 ms decoding). Code is available at \codelink{https://github.com/stcnjtech/DCVC-X-Intra}{\textcolor{blue}{link}}.%
\end{abstract}


\section{Introduction}
Neural image compression (NIC), driven by end-to-end rate-distortion optimization, has advanced rapidly and achieved performance competitive with or even superior to traditional codecs such as VVC\cite{balle2016end, balle2018variational, minnen2018joint, cheng2020learned, minnen2020channel}. In NIC, the entropy model estimates the probability distribution of quantized latents and plays a crucial role in determining the bitrate of the compressed bitstream. To improve entropy estimation, existing methods have developed various context models, including hyperprior-based models\cite{balle2018variational}, spatial\cite{minnen2018joint, cheng2020learned} or channel-wise\cite{minnen2020channel} autoregressive models, and joint spatial-channel context models\cite{li2024neural}. These approaches have sought a better balance between compression performance and coding efficiency.

However, existing context models primarily rely on internal priors from the current image, such as decoded latents and hyperprior features, while overlooking the common patterns and typical textures shared across natural images\cite{lu2025learnedimagecompressiondictionarybased}. Such dataset-level external priors contain rich global statistics and can complement distribution estimation when single-image information is insufficient\cite{oord2018neuraldiscreterepresentationlearning}, especially at low bitrates\cite{esser2021tamingtransformershighresolutionimage, jia2025generativelatentcodingultralow, qi2025generativelatentcodingultralow} or in complex texture regions\cite{lu2025learnedimagecompressiondictionarybased}. Exploiting these priors, however, raises two challenges: (i) internal and external priors differ substantially in representation, requiring efficient deep fusion without heavy computational overhead; and (ii) the distribution shift between them may even degrade entropy estimation if not properly constrained.

\begin{figure}[!t]
  \centering
  \includegraphics[width=0.93\linewidth]{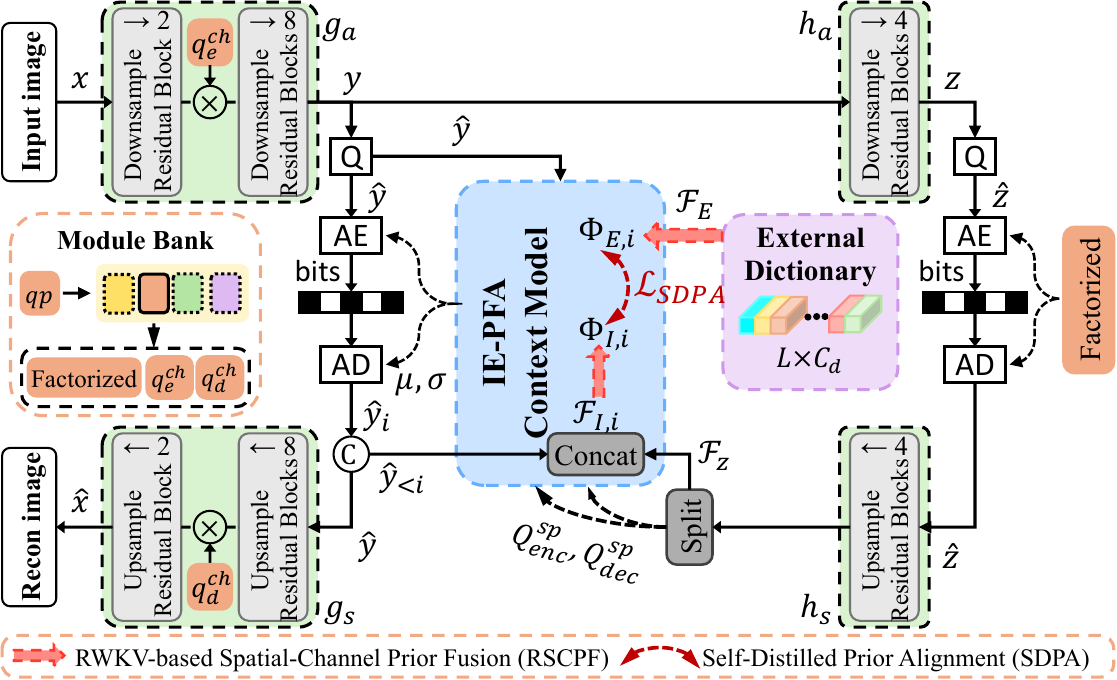}
  \caption{Overall architecture of our proposed model. The core is the Internal-External Prior Fusion-Alignment (IE-PFA) context model, which estimates the latent distribution by jointly exploiting internal priors and external priors provided by a learnable dictionary capturing common patterns of natural images.}
  \label{fig:architecture}
\end{figure}

\begin{figure*}[htb]
  \centering
  \includegraphics[width=\textwidth]{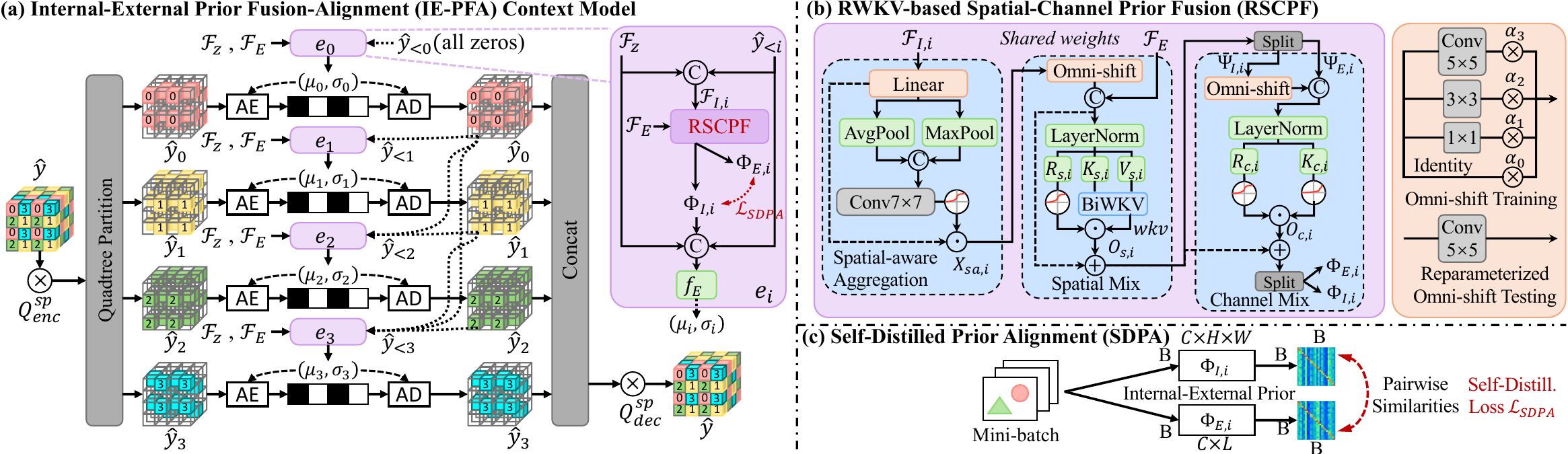}
  \caption{(a) The proposed IE-PFA context model with four branches based on quadtree spatial-channel joint modeling. Each branch $e_i$ mainly consists of (b) the RWKV-based Spatial-Channel Prior Fusion (RSCPF) module and (c) the Self-Distilled Prior Alignment (SDPA) mechanism.}
  \label{fig:iepfa}
\end{figure*}

To this end, we propose the Internal-External Prior Fusion-Alignment (IE-PFA) context model, which adopts a learnable dictionary shared between the encoder and decoder to provide external priors without transmitting any additional bits. Within a quadtree spatial-channel modeling framework, we design the RWKV-based Spatial-Channel Prior Fusion (RSCPF) module to efficiently fuse internal and external priors along both spatial and channel dimensions with linear complexity. To mitigate the distribution shift between the two priors, we further introduce a training-only Self-Distilled Prior Alignment (SDPA) mechanism, which uses mini-batch similarity structures as soft supervision to align the external prior with the internal one at zero inference cost. In addition, we use a three-stage progressive training strategy to further promote effective fusion and alignment of the two priors, and introduce a shared rate-adjustment module bank to enable smooth variable-rate adjustment within a single model. Our main contributions are:

\begin{itemize}
\item We propose IE-PFA, which introduces external priors via a shared learnable dictionary and jointly estimates entropy parameters with internal priors, exploiting cross-image statistics without extra bits.
\item We design the linear-complexity RSCPF fusion module and the training-only SDPA alignment mechanism, complemented by a three-stage progressive training strategy.
\item Experiments show that our method, with only 29.9M parameters, achieves an average BD-Rate saving of 21.65\% over VTM-22.0-Intra on seven benchmarks, outperforming advanced NIC methods while offering the fastest coding speed and smooth single-model variable-rate adaptation.
\end{itemize}

\section{Proposed Methods}
\subsection{Overview}

The overall architecture of the proposed codec, named DCVC-X-Intra, is illustrated in Fig.~\ref{fig:architecture}. The analysis transform $g_a$ maps an input image $x$ to the latent $y=g_a(x) \in \mathbb{R}^{C \times H \times W}$, which is quantized as $\hat{y}=Q(y)$ under estimated Gaussian parameters $(\mu, \sigma)$ and then losslessly entropy-coded; the synthesis transform $g_s$ reconstructs $\hat{x}=g_s(\hat{y})$ at the decoder. For entropy modeling, the hyper-analysis transform $h_a$ extracts side information $z=h_a(y)$, coded with a factorized prior after quantization, and the hyper-synthesis transform $h_s$ decodes $\hat{z}$ into the hyperprior feature $\mathcal{F}_z \in \mathbb{R}^{C \times H \times W}$ and the spatial-wise quantization steps $Q_{enc}^{sp}, Q_{dec}^{sp} \in \mathbb{R}^{1 \times H \times W}$.

The proposed IE-PFA is the core of the entropy model. Based on quadtree spatial-channel joint modeling, it partitions the latent into four branches coded sequentially, and estimates entropy parameters by jointly leveraging two types of priors: the \emph{internal} prior $\mathcal{F}_{I, i}$, formed by concatenating the decoded groups $\hat{y}_{<i}=\{\hat{y}_0, \cdots, \hat{y}_{i-1}\}$ with $\mathcal{F}_z$, models intra-image dependencies; the \emph{external} prior $\mathcal{F}_E \in \mathbb{R}^{L \times C_d}$ is a learnable dictionary shared by the encoder and decoder, capturing common patterns and typical textures of natural image datasets. The structural details of IE-PFA are presented in Sec.~\ref{sec:iepfa}. Moreover, unlike previous end-to-end schemes, we adopt a three-stage progressive training strategy (Sec.~\ref{sec:loss}) to fully fuse and align the two priors.

\subsection{Internal-External Prior Fusion-Alignment Context Model}
\label{sec:iepfa}

\begin{figure*}[t]
  \centering
  \includegraphics[width=\textwidth]{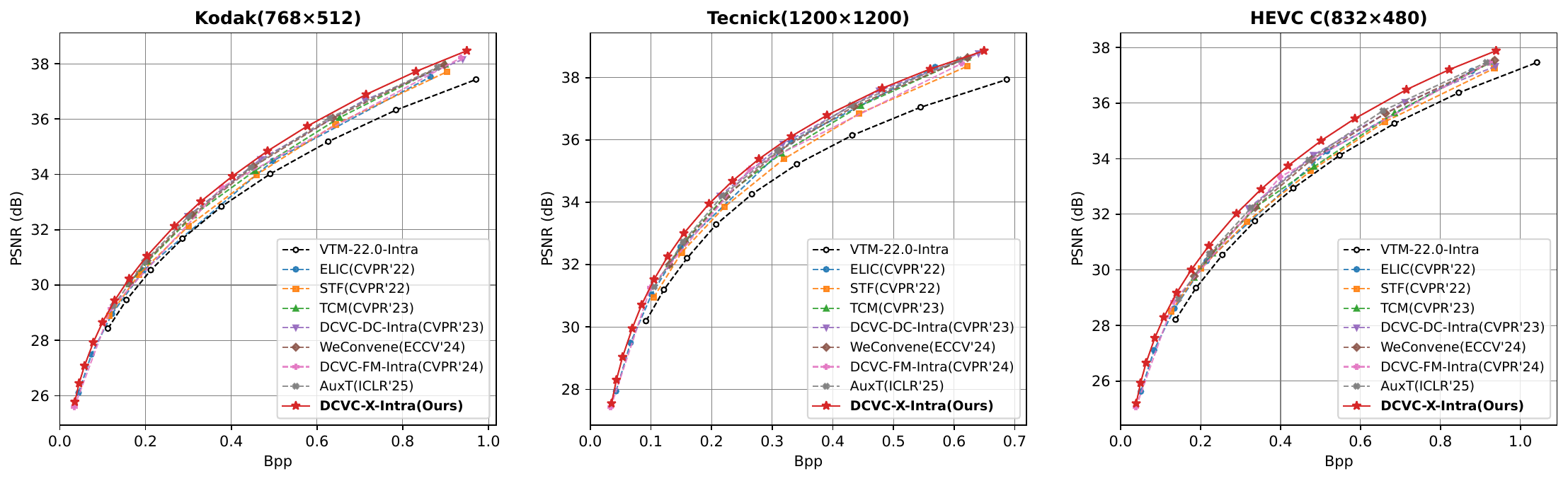}
  \caption{Rate-distortion curves of different methods on the Kodak, Tecnick, and HEVC C datasets.}
  \label{fig:rd}
\end{figure*}
Existing context models typically rely only on internal priors and fail to exploit dataset-level regularities\cite{he2022elic, liu2023learned, fu2024weconvene, li2025disentangled}. IE-PFA (Fig.~\ref{fig:iepfa}) introduces the external prior via a shared learnable dictionary: during training, it is aligned to the internal prior through a self-distillation loss, gradually learning the typical structures of natural images; at inference it in turn guides and refines the internal prior, providing a global complement when single-image information is insufficient. As shown in Fig.~\ref{fig:iepfa}(a), IE-PFA first multiplies $\hat{y}$ by $Q_{enc}^{sp}$ to modulate its spatial magnitude, then codes the four quadtree groups sequentially. The $i$-th branch $e_i$ takes $\mathcal{F}_{I, i}$ ($\hat{y}_{<i}$ is all-zero when $i=0$) and $\mathcal{F}_E$ as input to estimate the entropy parameters of $\hat{y}_i$:
\begin{equation}
  \mu_i,\sigma_i = e_i\left(\hat{y}_{<i}, \mathcal{F}_z, \mathcal{F}_E\right).
\end{equation}
Each branch comprises the RSCPF module, which interactively fuses $\mathcal{F}_{I, i}$ and $\mathcal{F}_E$ into the fused features $\Phi_{I, i}$ and $\Phi_{E, i}$, and the SDPA mechanism, which aligns the external fused feature to the internal one during training. The fused feature $\Phi_{I, i}$ is then merged with $\hat{y}_{<i}$ and $\mathcal{F}_z$ and fed into the entropy parameter module to predict more accurate $\mu_i$ and $\sigma_i$. Finally, all decoded groups are concatenated and multiplied by $Q_{dec}^{sp}$ to obtain the decoded latent.

\noindent \textbf{RSCPF Module.} As shown in Fig.~\ref{fig:iepfa}(b), RSCPF, which is shared across branches to greatly reduce parameters, contains spatial-aware aggregation, spatial mixing, and channel mixing. The spatial-aware aggregation employs a lightweight spatial attention to dynamically weight each position of the internal prior, $X_{sa, i}=SA(\mathcal{F}_{I, i})\odot \mathcal{F}_{I, i}$, capturing its typical textures. Inspired by linear-complexity attention\cite{gu2023mamba, dao2024transformers, peng-etal-2023-rwkv, zhai2021attentionfreetransformer, ICLR2025_ce65173b, yang2025restore}, the spatial mixing introduces the RWKV-based\cite{peng-etal-2023-rwkv} bidirectional WKV (BiWKV) attention to efficiently capture long-range dependencies and fuse the spatial features of the two priors: $X_{sa, i}$ first passes through an Omni-Shift\cite{yang2025restore} operation to capture 2D local context (reparameterized from multiple training-time kernels into a single kernel at inference), which is concatenated with $\mathcal{F}_E$ and projected via LayerNorm into the receptance $R_{s, i}$, key $K_{s, i}$, and value $V_{s, i}$. BiWKV then computes the global attention:
\begin{equation}
  wkv_t=\frac{\sum_{j=1,j\neq t}^{T}e^{-(|t-j|-1)/T\cdot w+k_j}\,v_j+e^{u+k_t}\,v_t}
             {\sum_{j=1,j\neq t}^{T}e^{-(|t-j|-1)/T\cdot w+k_j}+e^{u+k_t}},
\end{equation}
where $k_j$ and $v_j$ denote the key and value at position $j$; $wkv$ is modulated by the sigmoid gate $\sigma(R_{s,i})$ and then projected to the output $O_{s,i}$. The channel mixing analogously fuses channel-wise features. Its input $O_{s, i}$ is split into $\Psi_{I, i}$ and $\Psi_{E, i}$. After applying Omni-Shift to $\Psi_{I, i}$, the two branches are merged and projected to produce $R_{c, i}$ and $K_{c, i}$, while $V_{c, i}$ is obtained via squared ReLU. Gating $V_{c, i}$ by $\sigma(R_{c, i})$ then yields $O_{c, i}$, which is split into fused priors $\Phi_{I, i}\in\mathbb{R}^{C \times H \times W}$ and $\Phi_{E, i}\in\mathbb{R}^{C\times L}$.

\noindent \textbf{SDPA Mechanism.} Motivated by the observation in knowledge distillation and multimodal alignment\cite{tung2019similaritypreservingknowledgedistillation, 2021Learning} that similar semantics exhibit similar activation patterns, SDPA aligns the two priors via mini-batch similarity structures, as shown in Fig.~\ref{fig:iepfa}(c). For a training mini-batch of size $B$, we compute the L2-normalized pairwise similarities:
\begin{equation}
\left\{
\begin{aligned}
  G_{I,i} &= \mathrm{L2Norm}\!\left(\Phi_{I,i}\cdot\Phi_{I,i}^{T}\right), \\
  G_{E,i} &= \mathrm{L2Norm}\!\left(\Phi_{E,i}\cdot\Phi_{E,i}^{T}\right),
\end{aligned}
\right.
\end{equation}
where $G_{I,i}$ and $G_{E,i}$ are $B\times B$ matrices whose entries measure the activation similarity of image pairs. The alignment loss sums the discrepancies over the four branches:
\begin{equation}
  \mathcal{L}_{SDPA}=\frac{1}{B^2}\sum_{i=0}^{3}\left\|G_{I,i}-G_{E,i}\right\|_F^2,
\end{equation}
where $\|\cdot\|_F$ is the Frobenius norm. Requiring no extra teacher network, the mini-batch similarity structure serves as soft supervision that aligns the two priors at the feature-distribution level and mitigates their distribution shift. The SDPA mechanism operates only during training and adds no inference overhead.

\subsection{Overall Loss}
\label{sec:loss}
We adopt a three-stage progressive training strategy with the loss
\begin{equation}
  \mathcal{L}=\lambda\left(\|x-\hat{x}\|^2+\alpha\mathcal{L}_{SDPA}\right)+\mathcal{R}(\hat{y})+\mathcal{R}(\hat{z}),
\end{equation}
where $\mathcal{R}(\cdot)$ denotes the bitrate and $\lambda$ controls the rate-distortion trade-off. Stage 1 optimizes only the internal prior with the standard rate-distortion loss ($\alpha=0$). Stage 2 introduces the external prior: $g_a$ and $h_a$ are frozen, the remaining modules are trained, and the alignment term is added ($\alpha=0.2$) so that the external prior is optimized toward the internal one. Stage 3 unfreezes all modules and fine-tunes end-to-end with the same joint loss as Stage 2, further strengthening prior fusion and alignment.

\subsection{Other Optimizations}

To support smooth variable-rate adjustment in a single model rather than training one model per bitrate\cite{li2024neural}, we introduce a shared rate-adjustment module bank (see Fig.~\ref{fig:architecture}, left), which learns a series of factorized modules to model diverse distributions for different quantization levels $qp \in [0, 63]$. According to $qp$, the channel-wise quantization parameters $q_{e}^{ch}, q_{d}^{ch}$ drawn from the bank modulate the channel magnitude of the latent, while the hyperprior module generates the spatial-wise parameters $Q_{enc}^{sp}, Q_{dec}^{sp}$ for spatially adaptive granularity. During training, $qp$ is randomly sampled from the 64 levels, and the corresponding rate-distortion weight $\lambda$ is log-linearly interpolated within a predefined interval $[\lambda_{\min}, \lambda_{\max}]$. In addition, we develop fused CUDA operators to reduce GPU memory I/O overhead and accelerate inference.

\begin{figure}[t]
  \centering
  \includegraphics[width=\linewidth]{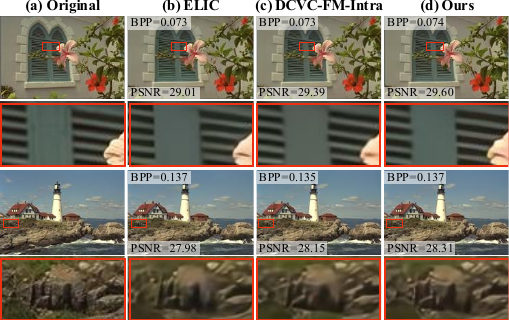}
  \caption{Visual comparison on Kodak images at low bitrates.}
  \label{fig:visual}
\end{figure}

\section{Experimental Results}
\label{sec:exp}

\subsection{Experimental Settings}
\label{ssec:settings}

\noindent \textbf{Datasets.}
Following previous works\cite{he2022elic, liu2023learned, li2025disentangled}, we train on the first 400K images of the ImageNet dataset\cite{deng2009imagenet} (shorter edge $\geq$ 256 pixels). Evaluation is conducted on Kodak\cite{franzen1999kodak}, CLIC Professional Validation\cite{george2020workshop}, Tecnick\cite{asuni2014testimages}, and the first 96 frames of each HEVC test sequence\cite{6316136} (Classes B--E).

\noindent \textbf{Baselines and Metrics.}
We compare with VVC\cite{9503377} (VTM-22.0-Intra: all-intra configuration, BT.709) and advanced neural codecs, including ELIC\cite{he2022elic}, STF\cite{zou2022devil}, TCM\cite{liu2023learned}, DCVC-DC-Intra\cite{li2023neural}, WeConvene\cite{fu2024weconvene}, DCVC-FM-Intra\cite{li2024neural}, and AuxT\cite{li2025disentangled}, using BD-Rate\cite{bjontegaard2001calculation} computed from RGB PSNR--bpp curves.

\begin{table*}[t]
  \centering
  \caption{BD-Rate (\%) relative to VTM-22.0-Intra on seven datasets. AVG is their average. GFLOPs are computed at the Kodak resolution ($768\times512$), and Enc./Dec. time (ms), including entropy coding, is averaged over the 24 Kodak images on an NVIDIA RTX 4090 GPU; $\dagger$ denotes our model with custom CUDA operators. The best and second-best results are in \textcolor{red}{\textbf{red bold}} and \second{blue}, respectively.}
  \label{tab:bdrate}
  \scriptsize
  \setlength{\tabcolsep}{4.0pt}
  \begin{tabular}{@{}lcccccccccccc@{}}
    \toprule
    \multirow{2}{*}{Method} & \multicolumn{8}{c}{BD-Rate over VTM-22.0-Intra (\%)} & \multirow{2}{*}{GFLOPs} & \multirow{2}{*}{Params} & Enc. & Dec. \\
    \cmidrule(lr){2-9}
     & Kodak & CLIC & Tecnick & HEVC B & HEVC C & HEVC D & HEVC E & AVG & & & (ms) & (ms) \\
    \midrule
    VTM-22.0-Intra & 0.00 & 0.00 & 0.00 & 0.00 & 0.00 & 0.00 & 0.00 & 0.00 & -- & -- & -- & -- \\
    ELIC\cite{he2022elic} & $-8.64$ & $-13.79$ & $-17.33$ & $-16.17$ & $-9.94$ & $-8.07$ & $-16.71$ & $-12.95$ & 371 & 41.9M & 91 & 119 \\
    STF\cite{zou2022devil} & $-8.22$ & $-10.01$ & $-11.00$ & $-13.97$ & $-6.67$ & $-2.12$ & $-14.96$ & $-9.56$ & 304 & 99.8M & 102 & 131 \\
    TCM\cite{liu2023learned} & $-11.70$ & $-17.00$ & $-17.03$ & $-16.70$ & $-8.06$ & $-5.67$ & $-15.26$ & $-13.06$ & 701 & 75.9M & 142 & 151 \\
    DCVC-DC-Intra\cite{li2023neural} & $-14.87$ & \second{$-18.75$} & \second{$-21.41$} & \textcolor{red}{$\mathbf{-23.54}$} & $-11.38$ & $-14.77$ & \textcolor{red}{$\mathbf{-29.77}$} & \second{$-19.21$} & 278 & 31.0M & 78 & 66 \\
    WeConvene\cite{fu2024weconvene} & $-13.37$ & $-16.77$ & $-17.73$ & $-17.42$ & $-10.52$ & $-8.29$ & $-16.79$ & $-14.41$ & 905 & 105.5M & 275 & 271 \\
    DCVC-FM-Intra\cite{li2024neural} & \second{$-15.02$} & $-17.60$ & $-19.94$ & $-19.39$ & \second{$-15.79$} & \second{$-18.67$} & $-25.91$ & $-18.90$ & \second{253} & \textcolor{red}{\textbf{26.6M}} & \second{71} & \second{62} \\
    AuxT\cite{li2025disentangled} & $-14.39$ & $-17.76$ & $-18.83$ & $-16.72$ & $-12.52$ & $-11.21$ & $-19.17$ & $-15.80$ & \textcolor{red}{\textbf{216}} & 46.2M & 170 & 166 \\
    \midrule
    DCVC-X-Intra(Ours) & \textcolor{red}{$\mathbf{-17.33}$} & \textcolor{red}{$\mathbf{-19.21}$} & \textcolor{red}{$\mathbf{-21.97}$} & \second{$-23.42$} & \textcolor{red}{$\mathbf{-19.01}$} & \textcolor{red}{$\mathbf{-21.98}$} & \second{$-28.60$} & \textcolor{red}{$\mathbf{-21.65}$} & 297 & \second{29.9M} & 74\,/\,\textcolor{red}{\textbf{61}}$^\dagger$ & 64\,/\,\textcolor{red}{\textbf{52}}$^\dagger$ \\
    \bottomrule
  \end{tabular}
\end{table*}

\noindent \textbf{Training Details.}
We train on 4 A100 GPUs, with $256 \times 256$ random crops, a batch size of 16 per GPU, and MSE as the distortion loss. A single model covering 64 quality levels with $\lambda \in [0.00067, 0.156]$ enables variable-rate coding. We adopt a three-stage training strategy: 100 epochs with a learning rate of $1\times10^{-4}$, followed by two stages of 50 epochs each with $1\times10^{-5}$.

\subsection{Rate-Distortion Performance and Complexity}
\label{ssec:rd}

As shown in Table~\ref{tab:bdrate}, among the compared methods, our method achieves the lowest BD-Rate on five of the seven datasets and the lowest average BD-Rate ($-21.65\%$). The rate-distortion curves on HEVC C in Fig.~\ref{fig:rd} show that our method outperforms the others at all tested bitrates, with a BD-Rate margin of 3.22 percentage points over the second-best DCVC-FM-Intra, suggesting that our prior modeling particularly benefits complex content. Fig.~\ref{fig:visual} compares the reconstructions of two Kodak images at low bitrates. At comparable bitrates (within 0.002~bpp), DCVC-X-Intra achieves the highest PSNR on both images: 29.60~dB versus 29.01~dB (ELIC) and 29.39~dB (DCVC-FM-Intra) on kodim07, and 28.31~dB versus 27.98~dB and 28.15~dB on kodim21. On the window shutter of kodim07, ELIC and DCVC-FM-Intra blur part of the thin slats, whereas DCVC-X-Intra keeps them sharp and clearly separated. On the rocks of kodim21, ELIC produces a smeared and discolored patch, and DCVC-FM-Intra over-smooths the surface so that the vertical cracks are largely lost, while DCVC-X-Intra retains these cracks.

Coding times in Table~\ref{tab:bdrate} are averaged over all 24 Kodak images. Without custom CUDA operators, our model (74~ms/64~ms) is the second fastest, behind only DCVC-FM-Intra (71~ms/62~ms). With custom CUDA operators, the encoding and decoding times are further reduced to 61~ms and 52~ms, i.e., by 17.6\% and 18.8\%, respectively, making our method the fastest. Moreover, with only 29.9M parameters, fewer than most competing methods, our model achieves a favorable performance-complexity trade-off.

\begin{table}[t]
  \centering
  \caption{Ablation study on Kodak: components added cumulatively and SDPA on DCAE\cite{lu2025learnedimagecompressiondictionarybased} (left), and dictionary size $L$ (right). BD-Rate (\%) is relative to VTM-22.0-Intra; latency (ms) is the total encoding and decoding time without custom CUDA operators.}
  \label{tab:ablation}
  {\scriptsize
  \setlength{\tabcolsep}{3pt}
  \begin{tabular*}{0.45\linewidth}[t]{@{\extracolsep{\fill}}lc@{}}
    \toprule
    Method & BD-Rate \\
    \midrule
    Baseline & $-13.43$ \\
    + RSCPF & $-15.84$ \\
    + Dictionary & $-15.97$ \\
    + SDPA & $\mathbf{-17.33}$ \\
    \midrule
    DCAE\cite{lu2025learnedimagecompressiondictionarybased} & $-21.24$ \\
    + SDPA & $\mathbf{-23.67}$ \\
    \bottomrule
  \end{tabular*}\hfill
  \begin{tabular*}{0.45\linewidth}[t]{@{\extracolsep{\fill}}ccc@{}}
    \toprule
    $L$ & BD-Rate & Latency \\
    \midrule
    0 & $-15.84$ & 124 \\
    32 & $-15.82$ & 135 \\
    64 & $-16.47$ & 136 \\
    128 & $\mathbf{-17.33}$ & 138 \\
    256 & $-17.27$ & 138 \\
    512 & $\mathbf{-17.33}$ & 142 \\[\dimexpr\aboverulesep+\lightrulewidth+\belowrulesep\relax]
    \bottomrule
  \end{tabular*}\par
  }
\end{table}

\begin{figure}[t]
  \centering
  \includegraphics[width=0.75\linewidth]{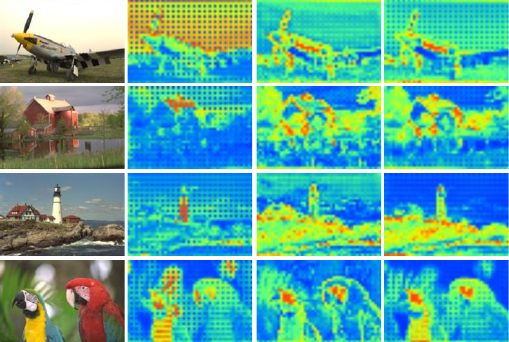}
  \caption{Visualization of priors on Kodak. From left to right: raw image $x$, original internal prior $\mathcal{F}_{I,i}$, unaligned fused prior $\Phi_{I,i}$, and SDPA-aligned fused prior $\Phi'_{I,i}$.}
  \label{fig:prior_vis}
\end{figure}

\subsection{Ablation Study}
\label{ssec:ablation}

\noindent \textbf{Effectiveness of RSCPF and SDPA.}
We build a baseline by removing all proposed modules and mechanisms (RSCPF, the external dictionary, and SDPA) and then add them back progressively until reaching the full DCVC-X-Intra. As shown in Table~\ref{tab:ablation} (left), adding RSCPF improves the BD-Rate from $-13.43\%$ to $-15.84\%$, and adding the dictionary without alignment brings only a marginal gain ($-15.97\%$); the external prior pays off once SDPA aligns it ($-17.33\%$). Fig.~\ref{fig:prior_vis} further visualizes the priors: the fused prior $\Phi_{I,i}$ responds more clearly to the main structures and textures, and the aligned prior $\Phi'_{I,i}$ focuses more on key structural and complex-texture regions, which qualitatively supports the joint effectiveness of RSCPF and SDPA for accurate entropy parameter estimation.

\noindent \textbf{Effect of the Dictionary Size.}
We further train models with dictionary sizes $L \in \{0, 32, 64, 128, 256, 512\}$ and report the BD-Rate and latency in Table~\ref{tab:ablation} (right). Despite a slight drop at $L=32$, the performance generally improves with $L$ and then saturates: the best BD-Rate is reached at $L=128$, whereas $L=256$ brings no gain at the same latency and $L=512$ only matches it with a higher latency (142~ms). Thus, $L=128$ achieves a good balance between compression performance and computational cost.

\noindent \textbf{Generalization of SDPA to DCAE.} DCAE\cite{lu2025learnedimagecompressiondictionarybased} also builds its entropy model on a learned dictionary, but its parameter count (119.2M) and encoding latency exceed those of all methods in Table~\ref{tab:bdrate}, so we evaluate it separately. Keeping its architecture unchanged, we add SDPA only as a training loss and retrain it with $\lambda \in \{0.05, 0.025, 0.013, 0.0067\}$, improving its BD-Rate from $-21.24\%$ to $-23.67\%$ (Table~\ref{tab:ablation}, left).

\section{Conclusion and Discussion}
In this paper, we propose a neural image compression model with an Internal-External Prior Fusion-Alignment (IE-PFA) context model, where an RWKV-based Spatial-Channel Prior Fusion (RSCPF) module fuses internal priors with a learnable external dictionary, and a Self-Distilled Prior Alignment (SDPA) mechanism aligns the two priors without extra inference cost. Together with a variable-rate design and custom CUDA operators, our model achieves state-of-the-art rate-distortion performance on seven benchmarks with only 29.9M parameters and the fastest coding speed among the compared methods. In future work, we will extend the external prior modeling to perceptual-oriented and video compression.

\clearpage
\bibliographystyle{IEEEtran}
\bibliography{reference}

\end{document}